\documentclass[10pt,twocolumn,letterpaper]{article}

\usepackage{iccv}
\usepackage{times}
\usepackage{epsfig}
\usepackage{graphicx}
\usepackage{amsmath}
\usepackage{amssymb}

\usepackage[square,numbers,sort&compress]{natbib}
\usepackage{enumitem}
\usepackage{algpseudocode}
\usepackage[dvipsnames]{xcolor}
\usepackage[font=small,labelfont=bf]{caption}
\usepackage{array}
\usepackage{multirow}
\usepackage{booktabs}
\usepackage{algorithm}
\usepackage{subcaption}
\usepackage[normalem]{ulem}
\usepackage{xparse}
\usepackage{pifont}
\usepackage{bm}
\usepackage{threeparttable}
\usepackage{etoolbox}

\usepackage{listings}
\usepackage{mwe} \usepackage{makecell}
\usepackage{color, colortbl}

\usepackage[scaled=0.85]{DejaVuSansMono}

\usepackage[breaklinks=true,bookmarks=false,colorlinks,bookmarks=false, citecolor=ForestGreen]{hyperref}
\usepackage[capitalize]{cleveref}

\crefname{section}{\S}{\S\S}
\crefname{subsection}{\S}{\S\S}
\crefformat{table}{Table~#2#1#3}
\crefformat{figure}{Fig~#2#1#3}
\crefformat{equation}{Eq~#2#1#3}

\preto\tabular{\setcounter{magicrownumbers}{0}}
\newcounter{magicrownumbers}

\newcommand{\N}{N}
\newcommand{\pN}{N'}

\newcommand{\bq}{\mathbf{q}}
\newcommand{\bqe}{\mathbf{q}^{e}}
\newcommand{\ba}{\mathbf{a}}
\newcommand{\bc}{\mathbf{c}}
\newcommand{\bo}{\mathbf{\Delta q}}
\newcommand{\bd}{\mathbf{d}}
\newcommand{\bs}{\mathbf{s}}
\newcommand{\bb}{\mathbf{b}}
\newcommand{\hbb}{\hat{\bb}}
\newcommand{\hbc}{\hat{\bc}}
\newcommand{\hba}{\hat{\ba}}
\newcommand{\hbd}{\hat{\bd}}
\newcommand{\hbs}{\hat{\bs}}
\newcommand{\bmatch}{\mathrm{match}}

\newcommand{\OURS}{3DETR\xspace}
\newcommand{\OURSm}{3DETR-m\xspace}

\newcommand{\sunrgbd}{SUN RGB-D\xspace}
\newcommand{\scannet}{ScanNetV2\xspace}

\newlength\savewidth

\newlength\thinwidth

\definecolor{Gray}{gray}{0.92}
\newcolumntype{a}{>{\columncolor{Gray}}c}
\definecolor{LightCyan}{rgb}{0.88,1,1}
\definecolor{altRowColor}{gray}{0.92}
\definecolor{highlightRowColor}{rgb}{0.95, 0.95, 1}
\newcommand{\colorrow}{\rowcolor{highlightRowColor}}

\newcommand{\rownumber}[1]{\textcolor{Cerulean}{#1}}

\iccvfinalcopy 
\def\iccvPaperID{1901} 

\ificcvfinal\fi

\begin{document}

\title{An End-to-End Transformer Model for 3D Object Detection
}

\author{
Ishan Misra \quad \quad Rohit Girdhar \quad \quad Armand Joulin \\
Facebook AI Research \\
{\small \url{https://facebookresearch.github.io/3detr}}
}

\maketitle
\ificcvfinal\thispagestyle{empty}\fi

\begin{abstract}

We propose 3DETR, an end-to-end Transformer based object detection model for 3D point clouds.
Compared to existing detection methods
that employ a number of 3D-specific inductive biases, 3DETR
requires minimal modifications to the vanilla Transformer block. Specifically, we find that a standard Transformer with non-parametric queries and Fourier positional embeddings is competitive with specialized architectures that employ libraries of 3D-specific operators with hand-tuned hyperparameters.
Nevertheless, 3DETR is conceptually simple and easy to implement,
enabling further improvements by incorporating 3D domain knowledge.
Through extensive experiments, we show 3DETR outperforms the well-established and highly optimized VoteNet baselines on the challenging ScanNetV2 dataset by 9.5\%.
Furthermore, we show 3DETR is applicable to 3D tasks beyond detection, and can serve as a building block for future research.

\end{abstract}

\section{Introduction}
\label{sec:intro}

3D object detection aims to identify and localize objects in 3D scenes. Such scenes,  often represented using {\em point clouds}, contain an unordered, sparse and irregular set of points captured using a depth scanner.
This set-like nature makes
point clouds significantly different from the traditional grid-like vision data like images and videos.
While there are other 3D representations such as multiple-views~\cite{su2015multi}, voxels~\cite{adams2010fast} or meshes~\cite{delaunay1934sphere}, they require additional post-processing to be constructed, and often loose information due to quantization.
Hence, point clouds have emerged as a popular 3D representation, and spurred the development of
specialized 3D architectures.
 
Many recent 3D detection models directly work on the 3D points to produce the bounding boxes.
Of particular interest, VoteNet~\cite{qi2019votenet} casts 3D detection as a set-to-set problem, \ie, transforming an unordered set of inputs (point cloud),
into an unordered set of outputs (bounding boxes).
VoteNet uses an encoder-decoder architecture:
the encoder is a PointNet++ network~\cite{qi2017pointnet} which converts the unordered point set into a unordered set of point features.
The point features are then input to a decoder that produces the 3D bounding boxes.
While effective, such architectures have required years of
careful development
by hand-encoding inductive biases, radii, and designing special 3D operators and loss functions.

In parallel to 3D, set-to-set encoder-decoder models have emerged as a competitive way to model 2D object detection.
In particular, the recent Transformer~\cite{vaswani2017attention} based model, called DETR~\cite{carion2020end}, casts 2D object detection as a set-to-set problem.
The self-attention operation in Transformers is designed to be permutation-invariant and capture long range contexts,
making them a natural candidate for processing unordered 3D point cloud data.
Inspired by this observation, we ask the following question: can we leverage
Transformers
to learn a 3D object detector without relying on hand-designed inductive biases?

\begin{figure}[!t]
\includegraphics[width=0.5\textwidth]{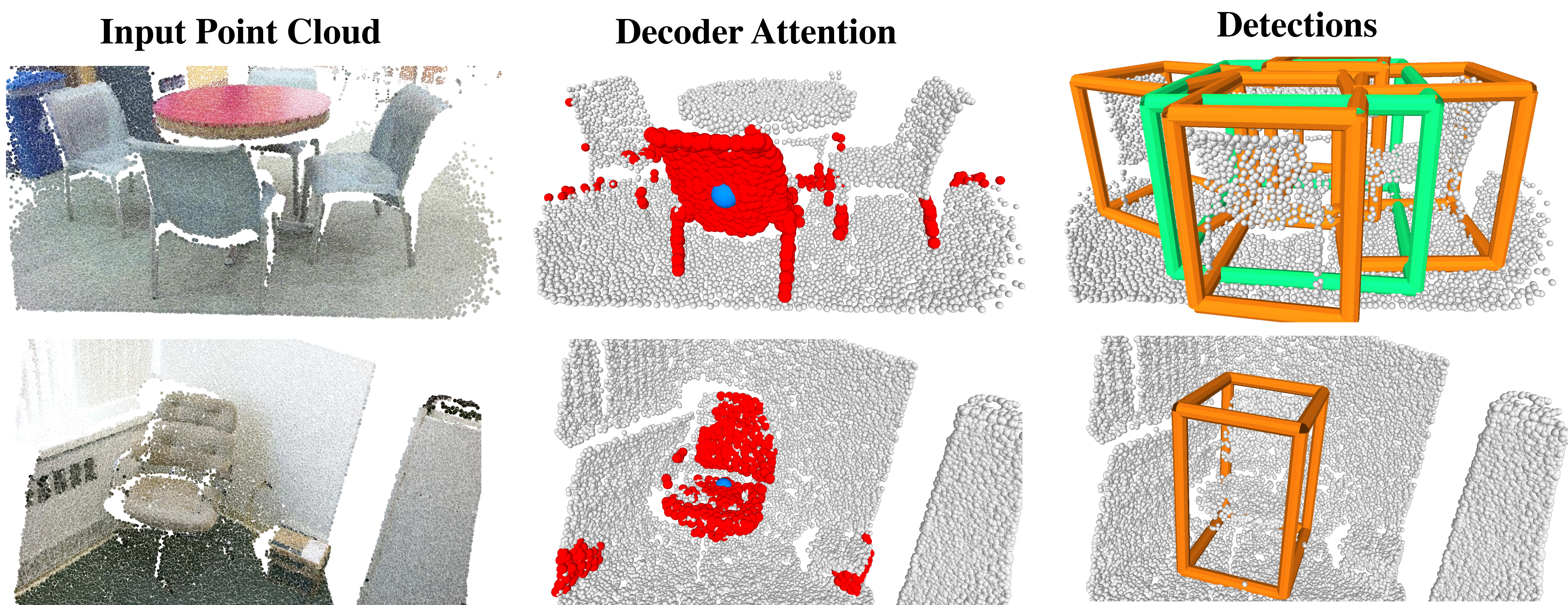}
\caption{\textbf{\OURS.} We train an end-to-end Transformer model for 3D object detection on point clouds.
Our model has a Transformer encoder for feature encoding and a Transformer decoder for predicting boxes.
For an unseen input, we compute the self-attention from the reference point (blue dot) to all points in the scene and display the points with the highest attention values in red.
The decoder attention groups points within an instance which presumably makes it easier to predict bounding boxes.
}
\label{fig:encoder_attn}
\end{figure}

To that end, we develop 3D DEtection TRansformer (\OURS) a simple to implement 3D detection method that uses fewer hand-coded design decisions and also casts detection as a set-to-set problem.
We explore the similarities between VoteNet and DETR, as well as between the core mechanisms of PointNet++ and the self-attention of Transformers to build our end-to-end Transformer-based detection model.
Our model follows the general encoder-decoder structure that is common to both DETR and VoteNet.
For the encoder, we replace the PointNet++ by a standard Transformer applied directly on the point clouds.
For the decoder, we consider the parallel decoding strategy from DETR with Transformer layers making two important changes to adapt it to 3D detection, namely non-parametric query embeddings and Fourier positional embeddings~\cite{tancik2020fourfeat}.

 \OURS removes many of the hard coded design decisions in VoteNet and PointNet++ while being simple to implement and understand.
Unlike DETR, \OURS does not employ a ConvNet backbone, and solely relies on Transformers trained from scratch.
Our transformer-based detection pipeline is flexible, and
as in VoteNet, any component can be replaced by other existing modules.
Finally, we show that 3D specific inductive biases can be easily incorporated in \OURS to further improve its performance.
On two standard indoor 3D detection benchmarks, \scannet and \sunrgbd we achieve 65.0\% AP and 59.0\% AP respectively, outperforming an improved VoteNet baseline by $9.5\%$ AP$_{50}$ on \scannet.

\begin{figure*}[!t]
\centering
\includegraphics[width=\textwidth]{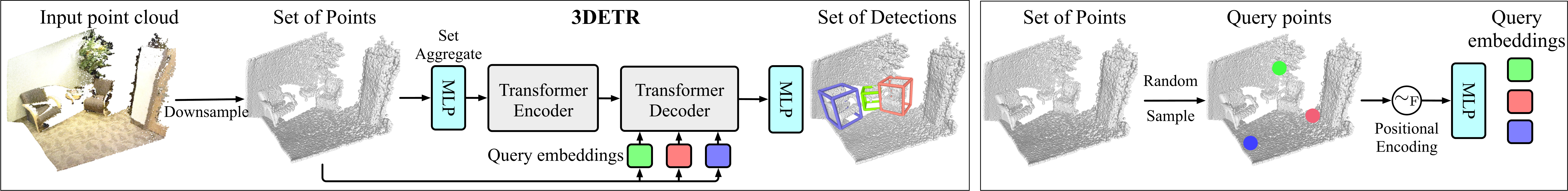}

\vspace{-0.1in}
\caption{\textbf{Approach.}
\emph{(Left)} \OURS is an end-to-end trainable Transformer that takes a set of 3D points (point cloud) as input and outputs a set of 3D bounding boxes.
The Transformer encoder produces a set of per-point features using multiple layers of self-attention.
The point features and a set of `query' embeddings are input to the Transformer decoder that produces a set of boxes.
We match the predicted boxes to the ground truth and optimize a set loss.
Our model does not use color information (used for visualization only).
\emph{(Right)}
We randomly sample a set of `query' points that are embedded and then converted into bounding box predictions by the decoder.
}
\label{fig:approach}
\end{figure*}
\vspace{-0.1in}
\section{Related Work}
\label{sec:related}
\vspace{-0.1in}

We propose a 3D object detection model composed of Transformer blocks.
We build upon prior work in 3D architectures, detection, and Transformers.

\vspace{0.02in}
\par \noindent \textbf{Grid-based 3D Architectures.}
Convolution networks can be applied to irregular 3D data after converting it into regular grids.
Projection methods~\cite{su2015multi,su2018splatnet,boulch2017unstructured,lawin2017deep,kanezaki2018rotationnet,lang2019pointpillars,tatarchenko2018tangent} project 3D data into 2D planes and convert it into 2D grids.
3D data can also be converted into a volumetric 3D grid by voxelization~\cite{tchapmi2017segcloud,riegler2017octnet,graham2015sparse,adams2010fast,hermosilla2018monte,li2018pointcnn,maturana2015voxnet,song2017semantic}.
We use 3D point clouds directly since they are suitable for \emph{set} based architectures such as the transformer.

\par \noindent \textbf{Point cloud Architectures.}
3D sensors often acquire data in the form of unordered point clouds.
When using unordered point clouds as input, it is desirable to obtain permutation invariant features.
Point-wise MLP based architectures~\cite{yang2019modeling,hu2020randla} such as PointNet~\cite{qi2017pointnet} and PointNet++~\cite{qi2017pointnet++} use permutation equivariant set aggregation (downsampling) and pointwise MLPs to learn effective representations. We use a single downsampling operation from~\cite{qi2017pointnet++} to keep the number of input points tractable in our model.

Graph-based models~\cite{wang2019graph,li2019deepgcns} can operate on unordered 3D data.
Graphs are constructed from 3D data in a variety of ways -- DGCNN~\cite{wang2019dynamic} and PointWeb~\cite{zhao2019pointweb} use local neighborhoods of points, SPG~\cite{landrieu2018large} uses attribute and context similarity and Jiang \etal~\cite{jiang2019hierarchical} use point-edge interactions.

Finally, continuous point convolution based architectures can also operate on point clouds.
The continuous weights can be defined using polynomial functions as in SpiderCNN~\cite{xu2018spidercnn} or linear functions as in Flex-Convolutions~\cite{groh2018flex}.
Convolutions can also be applied by soft-assignment matrices~\cite{verma2018feastnet} or specific ordering~\cite{li2018pointcnn}.
PointConv~\cite{wu2019pointconv} and KPConv~\cite{thomas2019kpconv} dynamically generate convolutional weights based on the input point coordinates, while InterpCNN~\cite{mao2019interpolated} uses these coordinates to interpolate weights.
We build upon the Transformer~\cite{vaswani2017attention} which is applicable for sets but not tailored for 3D.

\par \noindent \textbf{3D Object Detection} is a well studied research area where methods predict three dimensional bounding boxes from 3D input data~\cite{qi2018frustum,pham2016geometrically,lahoud20193d,Song2014SlidingSF,Song_2016_CVPR,zhu2020ssn,simony2018complex,vora2020pointpainting,wang2015voting}.
Many methods avoid expensive 3D operations by using 2D projection.
MV3D~\cite{chen2017multi}, VoxelNet~\cite{zhou2018voxelnet} use a combination of 3D and 2D convolutions.
Yan \etal~\cite{yan2018second} simplify the 3D operation while~\cite{yang2018pixor} uses a 2D projection, and~\cite{wang2020pillar} uses `pillars' of voxels.
We focus on methods that directly use 3D point clouds~\cite{yi2019gspn,shi2019pointrcnn,pham2019jsis3d,wang2019associatively}.
PointRCNN~\cite{shi2019pointrcnn} and PVRCNN~\cite{shi2020pv} are 2-stage detection pipelines similar to the popular R-CNN framework~\cite{Ren15fasterrcnn} for 2D images.
While these methods are related to our work, for simplicity we build a single stage detection model as done in~\cite{qi2019votenet,yang20203dssd,engelmann20203d,gwak2020gsdn}.
VoteNet~\cite{qi2019votenet} uses Hough Voting on sparse point cloud inputs and detects boxes by feature sampling, grouping and voting operations designed for 3D data.
VoteNet is a building block for many follow up works.
3D-MPA~\cite{engelmann20203d} combines voting with a graph ConvNet for refining object proposals and uses specially designed 3D geometric features for aggregating detections.
HGNet~\cite{chen2020hierarchical} improves Hough Voting and uses a hierarchical graph network with feature pyramids.
H3DNet~\cite{zhang2020h3dnet} improves VoteNet by predicting 3D primitives and uses a geometric loss function.
We propose a simple detection method that can serve as a building block for such innovations in 3D detection.
\par \noindent \textbf{Transformers in Vision.} The Transformer architecture by Vaswani \etal~\cite{vaswani2017attention} has been immensely successful across domains like NLP~\cite{radford2018improving,devlin2018bert}, speech recognition~\cite{luscher2019rwth,synnaeve2019end}, image recognition~\cite{parmar2018image,carion2020end,wang2018non,hu2018relation,dosovitskiy2020image}, and for cross-domain applications~\cite{lu2019vilbert,su2019vl,tan2019lxmert}.
Transformers are well suited for operating on 3D points since they are naturally permutation invariant.
Attention based methods have been used for building 3D point representations for retrieval~\cite{zhang2019pcan}, outdoor 3D detection~\cite{liu2020tanet,yin2020lidar,paigwar2019attentional}, object classification~\cite{yang2019modeling}.
Concurrent work~\cite{zhao2020point,pan20203d} also uses the Transformer architecture for 3D.
While these methods use 3D specific information to modify the Transformer, we push the limits of the standard Transformer.
Our work is inspired by the recent DETR model~\cite{carion2020end} for object detection in images by Carion \etal~\cite{carion2020end}.
Different from Carion \etal, our model is an end-to-end transformer (no convolutional backbone) that can be trained from scratch and has important design differences such as non-parametric queries to enable 3D detection.

\vspace{-0.05in}
\section{Approach}
\label{sec:approach}
\vspace{-0.05in}

We briefly review prior work in 3D detection and their conceptual similarities to \OURS.
Next, we describe \OURS,
simplifications in bounding box parametrization and the simpler set-to-set objective function.

\vspace{-0.05in}
\subsection{Preliminaries}
\label{sec:preliminaries}
\vspace{-0.05in}

The recent VoteNet~\cite{qi2019votenet} framework forms the basis for many detection models in 3D, and like our method, is a set-to-set prediction framework. VoteNet uses a specialized 3D encoder and decoder architecture for detection. It combines these models with a Hough Voting loss designed for sparse point clouds.
The encoder is a PointNet++~\cite{qi2017pointnet++} model that uses a combination of multiple downsampling (set-aggregation) and upsampling (feature-propagation) operations that are specifically designed for 3D point clouds.
The VoteNet ``decoder'' predicts bounding boxes in three steps - 1) each point `votes' for the center coordinate of a box; 2) votes are aggregated within a fixed radius to obtain `centers'; 3) bounding boxes are predicted around `centers'.
\emph{BoxNet}~\cite{qi2019votenet} is a non-voting alternative to VoteNet that randomly samples `seed' points from the input and treats them as `centers'.
However, BoxNet achieves much worse performance than VoteNet as the voting captures additional context in sparse point clouds and yields better `center' points.
As noted by the authors~\cite{qi2019votenet}, the multiple hand-encoded radii used in the encoder, decoder, and the loss function are important for detection performance and have been carefully tuned~\cite{qi2017pointnet++,qi2017pointnet}.

The Transformer~\cite{vaswani2017attention} is a generic architecture that can work on set inputs and capture large contexts by computing self-attention between all pairs of input points.
Both these properties make it a good candidate model for 3D point clouds.
Next, we present our \OURS model which uses a Transformer for both the encoder and decoder with minimal modifications and has minimal hand-coded information for 3D.
\OURS uses a simpler training and inference procedure.
We also highlight similarities and differences to the DETR model for 2D detection.

\vspace{-0.05in}
\subsection{\OURS: Encoder-decoder Transformer}
\label{sec:pointdetr}
\vspace{-0.05in}

\OURS takes as input a 3D point cloud and predicts the positions of objects in the form of 3D bounding boxes.
A point cloud is a unordered set of $N$ points where each point is associated with its $3$-dimensional XYZ coordinates.
The number of points is very large and we use the set-aggregation downsampling operation from~\cite{qi2017pointnet++} to downsample the points and project them to $\pN$ dimensional features.
The resulting subset of $\pN$ features is passed through an encoder to also obtain a set of $\pN$ features.
A decoder takes these features as input and predicts multiple bounding boxes using a parallel decoding scheme inspired by~\cite{carion2020end}.
Both encoder and decoder use standard Transformer blocks with `pre-norm'~\cite{klein2017opennmt} and we refer the reader to Vaswani~\etal~\cite{vaswani2017attention} for details.
~\cref{fig:approach} illustrates our model.
\vspace{0.02in}
\par \noindent \textbf{Encoder.}
The downsample and set-aggregation steps provide a set of $\pN$ features of $d=256$ dimensions using an MLP with two hidden layers of $64, 128$ dimensions.
The set of $\pN$ features is then passed to a Transformer to also produce a set of $N'$ features of $d\!=\!256$ dimensions.
The Transformer applies multiple layers of self-attention and non-linear projections.
We do not use downsampling operations in the Transformer, and use the standard self-attention formulation~\cite{vaswani2017attention}.
Thus, the Transformer encoder has no specific modifications for 3D data.
We omit positional embeddings of the coordinates from the encoder since the input already contains information about the XYZ coordinates.

\vspace{0.02in}
\par \noindent \textbf{Decoder.}
Following Carion \etal~\cite{carion2020end}, we frame detection as a set prediction problem, \ie, we simultaneously predict a set of boxes with no particular ordering.
This is achieved with a parallel decoder composed of Transformer blocks.
This decoder takes as input the $\pN$ point features and a set of $B$ query embeddings $\{\bqe_1,\dots,\bqe_B\}$ to produce a set of $B$ features that are then used to predict 3D-bounding boxes.
In our framework, the query embeddings $\bqe$ represent locations in 3D space around which our final 3D bounding boxes are predicted.
We use positional embeddings in the decoder as it does not have direct access to the coordinates (operates on encoder features and query embeddings).

\vspace{0.02in}
\par \noindent \textbf{Non-parametric query embeddings.}
Inspired by seed points used in VoteNet and BoxNet~\cite{qi2019votenet}, we use non-parametric embeddings computed from `seed' XYZ locations.
We sample a set of $B$ `query' points $\{ \bq_{i} \}_{i=1}^{B}$ randomly from the $\pN$ input points (see~\cref{fig:approach}).
We use Farthest Point Sampling~\cite{qi2017pointnet++} for the random samples as it
ensures a good coverage of the original set of points.
We associate each query point $\bq_i$ with a query embedding $\bqe_i$, by converting the coordinates of $\bq_i$ into Fourier positional embeddings~\cite{tancik2020fourfeat} followed by projection with a MLP.

\vspace{0.05in}
\par \noindent \textbf{\OURSm: Inductive biases into \OURS.}
As a proof of concept that our model is flexible, we modify our encoder to include inductive biases in 3D data, while keeping the decoder and loss fixed.
We leverage a weak inductive bias inspired by PointNet++,
\ie, local feature aggregation matters more than global aggregation.
Such an inductive bias can be easily implemented in Transformers by applying a mask to the self-attention~\cite{vaswani2017attention}.
The resulting model, \OURS-\emph{m} has a \emph{m}asked self-attention encoder with the same decoder and loss function as \OURS.
\OURSm uses a three layer encoder which has an additional downsampling operation (from $\pN\!=\!2048$ to $N''\!=\!1024$ points) after the first layer.
Every encoder layer applies a binary mask of $N''\times N''$ to the self-attention operation.
Row $i$ in the mask indicates which of the $N''$ points lie within the $\ell_{2}$ radius of point $i$.
We use the radius values of $[0.16,0.64,1.44]$.
Compared to PointNet++, \OURSm does not rely on multiple layers of 3D feature aggregation and 3D upsampling.

\subsection{Bounding box parametrization and prediction}
\label{sec:bounding_box_param}
The encoder-decoder architecture produces a set of $B$ features, that are fed into prediction MLPs to predict bounding boxes.
A 3D bounding box has the attributes
\textbf{(a)} its location, \textbf{(b)} size, \textbf{(c)} orientation, and \textbf{(d)} the class of the object contained in it.
We describe the parametrization of these attributes and their associated prediction problems.

The prediction MLPs produce a box around every query coordinate $\bq$.
\noindent\textbf{(a) Location:} We use the XYZ coordinates of box's center $\bc$.
We predict this in terms of an offset $\bo$ that is added to the query coordinates, \ie, $\bc = \bq + \bo$.

\noindent\textbf{(b) Size:} Every box is a 3D rectangle and we define its size around the center coordinate $\bc$ using XYZ dimensions $\bd$.

\noindent\textbf{(c) Orientation:} In some settings~\cite{song2015sun}, we must predict the orientation of the box, \ie, the angle it forms compared to a given referential.
We follow~\cite{qi2019votenet} and quantize the angles into $12$ bins from $[0, 2\pi)$ and note the quantization residual.
Angular prediction involves predicting the the quantized `class' of the angle and the residual to obtain the continuous angle $a$.
\noindent\textbf{(d) Semantic Class:} We use a one-hot vector $\bs$ to encode the object class contained in the bounding box.
We include a `background' or `not an object' class as some of the predicted boxes may not contain an object.

Putting together the attributes of a box, we have two quantities: the predicted boxes $\hbb$ and the ground truth boxes $\bb$.
Each predicted box $\hbb = [ \hbc, \hbd, \hba, \hbs ]$ consists of (1) geometric terms $\hbc, \hbd \in [0,1]^{3}$ that define the box center and dimensions respectively, $\hba = [\hba_c, \hba_r ]$ that defines the quantized class and residual for the angle; (2) semantic term $\hbs = [0, 1]^{K+1}$ that contains the probability distribution over the $K$ semantic object classes and the `background' class.
The ground truth boxes $\bb$ also have the same terms.

\subsection{Set Matching and Loss Function}
\label{sec:set_loss}
To train the model, we first match the set of $B$ predicted 3D bounding boxes $\{\hbb\}$ to the ground truth bounding boxes $\{\bb\}$.
While VoteNet uses hand-defined radii to do such set matching, we follow~\cite{carion2020end} to perform a bipartite graph matching which is simpler, generic (see~\cref{sec:components}) and robust to Non-Maximal Suppression.
We compute a loss for each predicted box using its matched ground truth box.

\par \noindent \textbf{Bipartite Matching.} We define a matching cost for a pair of boxes, predicted box $\hbb$ and ground truth box $\bb$, using a geometric and a semantic term.

\begin{multline}
C_{\bmatch}(\hbb, \bb) =  \underbrace{- \lambda_{1} \mathrm{GIoU}(\hbb, \bb) + \lambda_{2} \|\hbc - \bc\|_{1}}_\text{geometric} \\
- \underbrace{\lambda_{3} \hbs[s_{\mathrm{gt}}] + \lambda_{4} (1 - \hbs[s_{\mathrm{bg}}]) }_\text{semantic} \\
\end{multline}

These terms are similar to the loss functions used for training the model and $\lambda$s are scalars used for a weighted combination.
The geometric cost measures the box overlap using GIoU~\cite{Rezatofighi_2018_CVPR} and the distance between the centers of the boxes. Box overlap automatically accounts for the box dimensions, angular rotation and is scale invariant.
The semantic cost measures the likelihood of the ground truth class $s_{\mathrm{gt}}$ under the predicted distribution $\hbs$ and the likelihood of the box features belonging to a foreground class, \ie, of not belonging to the background class $s_{\mathrm{bg}}$.

We compute the optimal bipartite matching between all the predicted boxes $\{\hbb\}$ and ground truth boxes $\{\bb\}$ using the Hungarian algorithm~\cite{kuhn1955hungarian} as in prior work~\cite{stewart2016end,carion2020end}.
As we predict a larger number of boxes than the ground truth, the predicted boxes that do not get matched are considered matched to the `background' class.
This encourages the model to not over-predict, a property that helps our model be robust to Non-Maximal Suppression (see~\cref{sec:ablations}).

\par \noindent \textbf{Loss function.} We use $\ell_{1}$ regression losses for the center and box dimensions, normalizing them both in the range $[0, 1]$ for scale invariance.
We use Huber regression loss for the angular residuals and cross-entropy losses for the angular classification and semantic classification.
\begin{multline}
\mathcal{L}_{\mathrm{\OURS}} = \lambda_{c} \|\hbc - \bc\|_{1} + \lambda_{d} \|\hbd - \bd\|_{1} + \lambda_{ar} \|\hba_r - \ba_r\|_{\mathrm{huber}} \\
-\lambda_{ac} \ba_c^\intercal \log \hba_c -\lambda_{s} \bs_c^\intercal \log \hbs_c
\end{multline}

Our final loss function is a weighted combination of the above five terms and we provide the full details in the appendix.
For predicted boxes matched to the `background' class, we only compute the semantic classification loss with the background class ground truth label.
For datasets with axis-aligned 3D bounding boxes, we also use a loss directly on the GIoU as in~\cite{carion2020end,Rezatofighi_2018_CVPR}.
We do not use the GIoU loss for oriented 3D bounding boxes as it is computationally involved.
\par \noindent \textbf{Intermediate decoder layers.} At training time, we use the same bounding box prediction MLPs to predict bounding boxes at every layer in the decoder.
We compute the set loss for each layer independently and sum all the losses to train the model.
At test time, we only use the bounding boxes predicted from the last decoder layer.

 \vspace{-0.05in}
\subsection{Implementation Details}
\vspace{-0.05in}
We implement \OURS using PyTorch~\cite{NEURIPS2019_9015} and use the standard \texttt{nn.MultiHeadAttention} module to implement the Transformer.
We use a single set aggregation operation~\cite{qi2017pointnet++} to subsample $\N'\!=\!2048$ points and obtain $256$ dimensional point features.
The \OURS encoder has 3 layers where each layer uses multiheaded attention with four heads and a two layer MLP with a `bottleneck' of $128$ hidden dimensions.
The \OURS decoder has 8 layers and closely follows the encoder, except that the MLP hidden dimensions are $256$.
We use Fourier positional encodings~\cite{tancik2020fourfeat} of the XYZ coordinates in the decoder.
The bounding box prediction MLPs are two layer MLPs with a hidden dimension of $256$.
Full architecture details in the appendix~\cref{sec:supp_architecture}.

All the MLPs and self-attention modules in the model use a dropout~\cite{srivastava2014dropout} of $0.1$ except in the decoder where we use a higher dropout of $0.3$.
\OURS is optimized using the AdamW optimizer~\cite{loshchilov2017decoupled} with the learning rate decayed by a cosine learning rate schedule~\cite{loshchilov2016sgdr} to $10^{-6}$, a weight decay of $0.1$, and gradient clipping at an $\ell_{2}$ norm of $0.1$.
We train the model on a single V100 GPU with a batchsize of $8$ for 1080 epochs.
We use the RandomCuboid augmentation from~\cite{zhang_depth_contrast} which reduces overfitting.

\vspace{-0.05in}
\section{Experiments}
\label{sec:experiments}
\vspace{-0.05in}

\par \noindent \textbf{Dataset and metrics.}
We evaluate models on two standard 3D indoor detection benchmarks - \scannet~\cite{Dai_2017_CVPR_scannet} and \sunrgbd-v1~\cite{song2015sun}.
\sunrgbd has 5K single-view RGB-D training samples with oriented bounding box annotations for 37 object categories.
\scannet has 1.2K training samples (reconstructed meshes converted to point clouds) with axis-aligned bounding box labels for 18 object categories.
For both datasets, we follow the experimental protocol from~\cite{qi2019votenet}:
we report the detection performance on the val set using mean Average Precision (mAP) at two different IoU thresholds of $0.25$ and $0.5$, denoted as AP$_{25}$ and AP$_{50}$.
Along with the metric, their protocol evaluates on the 10 most frequent categories for \sunrgbd.

\subsection{\OURS on 3D Detection}
\label{sec:comparison}

\begin{table}[!t]
          \centering
    \begin{tabular}{@{}l|cccc@{}}
    \toprule
    \textbf{Method} & \multicolumn{2}{c}{\textbf{\scannet}} & \multicolumn{2}{c}{\textbf{\sunrgbd}}\\
    & AP$_{25}$ & AP$_{50}$ & AP$_{25}$ & AP$_{50}$ \\
    \hline
    BoxNet$^\dagger$~\cite{qi2019votenet} & 49.0 & 21.1 & 52.4 & 25.1 \\
    \colorrow \OURS & 62.7 & 37.5 & 58.0 & 30.3 \\
\hline
    VoteNet$^\dagger$~\cite{qi2019votenet} & 60.4 & 37.5 & 58.3 & 33.4 \\
    \colorrow \OURSm & 65.0 & 47.0 & 59.1 & 32.7 \\
\hline
    H3DNet~\cite{zhang2020h3dnet} & 67.2 & 48.1 & 60.1 & 39.0 \\
    \bottomrule
                \end{tabular}
\vspace{-0.1in}
\caption{\textbf{Evaluating \OURS on 3D detection.}
We compare \OURS with BoxNet and VoteNet methods and denote by $^\dagger$ our improved implementation of these baselines.
\OURS achieves comparable or better performance to these improved baselines despite having fewer hand-coded 3D or detection specific decisions.
We report state-of-the-art performance from~\cite{zhang2020h3dnet} that improves VoteNet by using 3D primitives. Detailed state-of-the-art comparison in~\cref{sec:supp_experiments}.
}
\label{tab:comparison}
\end{table}

In this set of experiments, we validate \OURS for 3D detection.
We compare it to the BoxNet and VoteNet models since they are conceptually similar to \OURS and are the foundations of many recent detection models.
For fair comparison, we use our own implementation of these models with the same optimization improvements used in \OURS -- leading to a boost of +2-4\% AP over the original paper (details in supplemental).
We also compare against a state-of-the-art method H3DNet~\cite{zhang2020h3dnet} and provide a more detailed comparison against other recent methods in the appendix.
\OURS models use $256$ and $128$ queries for \scannet and \sunrgbd datasets.
\begin{figure*}[!t]
\centering
\includegraphics[width=.9\textwidth]{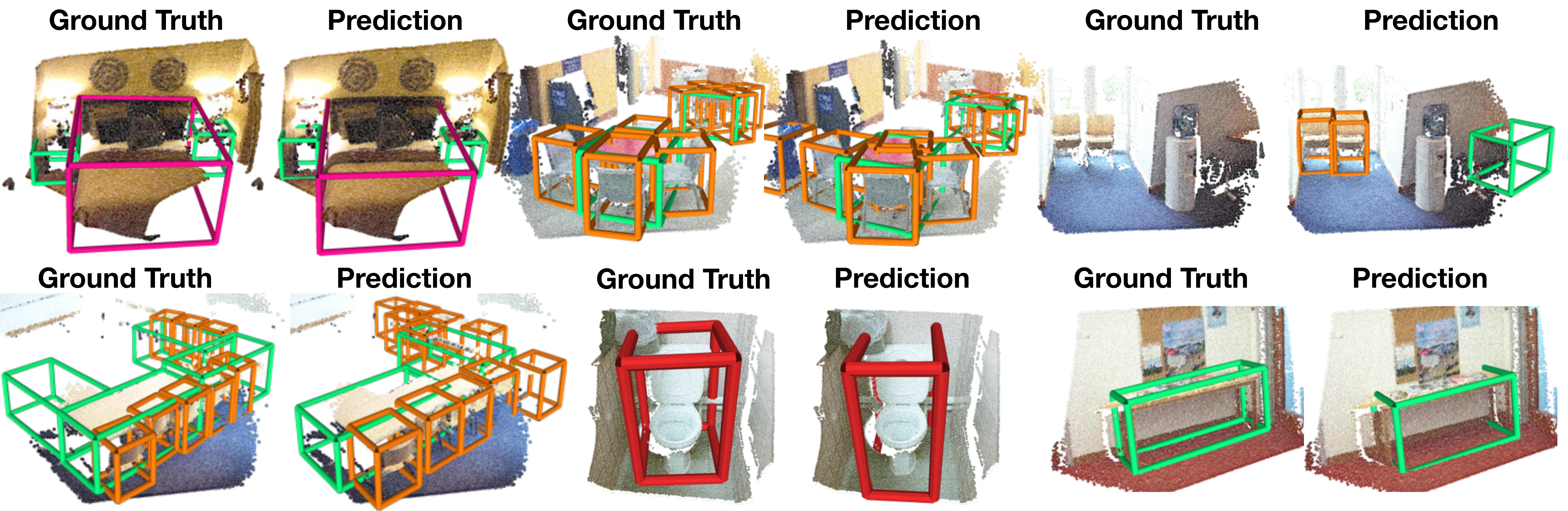}
\vspace{-0.1in}
\caption{\textbf{Qualitative Results using \OURS.} Detection results for scenes from the val set of the \sunrgbd dataset.
\OURS does not use color information (used only for visualization) and predicts boxes from point clouds.
\OURS can detect objects even with single-view depth scans and predicts amodal boxes \eg, the full extent of the bed (top left) including objects missing in the ground truth (top right).
\label{fig:detection_results}
}
\end{figure*}

\vspace{0.05in}
\par \noindent \textbf{Observations.}
We summarize results in~\cref{tab:comparison}.
The comparison between BoxNet and \OURS is particularly relevant since both methods predict boxes around location queries while VoteNet uses 3D Hough Voting to obtain queries.
Our method significantly outperforms BoxNet on both the datasets with a gain of $+13\%$ AP$_{25}$ on \scannet and $+3.9\%$ AP$_{25}$ on \sunrgbd.
Even when compared with VoteNet, our model achieves competitive performance, with $+2.3\%$ AP$_{25}$ on \scannet and $-1.5\%$ AP$_{25}$ on \sunrgbd.
\OURSm, which uses the masked Transformer encoder, achieves comparable performance to VoteNet on \sunrgbd and a gain of $+4.6\%$ AP$_{25}$ and $+9.5\%$ AP$_{50}$ on \scannet.

Compared to a state-of-the-art method, H3DNet~\cite{zhang2020h3dnet}, that builds upon VoteNet, \OURSm is within a couple of AP$_{25}$ points on both datasets (more detailed comparison in~\cref{sec:supp_experiments}).
These experiments validate that a encoder-decoder detection model based on the standard Transformer is competitive with similar models tailored for 3D data.
Just as the VoteNet model was improved by the innovations of H3DNet~\cite{zhang2020h3dnet}, HGNet~\cite{chen2020hierarchical}, 3D-MPA~\cite{engelmann20203d}, similar innovations could be integrated to our model in the future.

\vspace{0.05in}
\par \noindent \textbf{Qualitative Results.} In~\cref{fig:detection_results}, we visualize a few detections and ground truth boxes from \sunrgbd.
\OURS detects boxes despite the partial (single-view) depth scans and also predicts amodal bounding boxes or missing annotations on \sunrgbd.

\subsection{Analyzing \OURS}
\label{sec:analysis}

We conduct a series of experiments to understand \OURS.
In~\cref{sec:components}, we explore the similarities between \OURS, VoteNet and BoxNet.
Next, in~\cref{sec:design_transformer}, we compare the design decisions in \OURS that enable 3D detection to the original components in DETR.

\vspace{-0.1in}
\subsubsection{Modules of VoteNet and BoxNet vs. \OURS}
\label{sec:components}
\vspace{-0.1in}
The encoder-decoder paradigm is flexible and we can test if the different modules in VoteNet, BoxNet and \OURS are interchangeable.
We focus on the encoders, decoders and losses and report the detection performance in~\cref{tab:components_encoder_loss,tab:components_decoder_loss}.
For simplicity, we denote the decoders and the losses used in BoxNet and VoteNet as Box and Vote respectively.
We use PointNet++ to refer to the modified PointNet++ architecture used in VoteNet~\cite{qi2019votenet}.

\begin{table}[!t]
    \mbox{}\hfill
      \centering
\setlength{\tabcolsep}{0.3em}\resizebox{\linewidth}{!}{
    \begin{threeparttable}
    \begin{tabular}{@{}ll|lll|cc|cc@{}}
    \toprule
    & \textbf{Method} & \textbf{Encoder} & \textbf{Decoder} & \textbf{Loss}  & \multicolumn{2}{c}{\textbf{\scannet}} & \multicolumn{2}{c}{\textbf{\sunrgbd}}\\
    &&&&& AP$_{25}$ & AP$_{50}$ & AP$_{25}$ & AP$_{50}$ \\
    \hline
        \colorrow &\OURS & Tx. & Tx. & Set & 62.7 & 37.5 & 58.0 & 30.3 \\
                        && PN++ & Tx.  & Set  & 61.4 & 34.7 & 56.8 & 26.9 \\
    \bottomrule
    \end{tabular}
    \begin{tablenotes}
    \item {\small \hspace{-0.25in} PN++: PointNet++~\cite{qi2017pointnet++}, Tx.: Transformer, Set loss~\cref{sec:set_loss}}
    \end{tablenotes}
    \end{threeparttable}
}
\vspace{-0.1in}
\caption{\textbf{\OURS with different encoders.}
We vary the encoder used in \OURS and observe that the performance is unchanged or slightly worse when moving to a PointNet++ encoder.
This suggests that the decoder design and the loss function in \OURS are compatible with prior 3D specific encoders.
}
\label{tab:components_encoder_loss}
\end{table}

\vspace{0.02in}
\par \noindent \textbf{Replacing the encoder.}
We train \OURS with a PointNet++ encoder (\cref{tab:components_encoder_loss}) and observe that the detection performance is unchanged or slightly worse compared to \OURS with a transformer encoder.
This shows that the design decisions in \OURS are broadly compatible with prior work, and can be used for designing better encoder models.

\vspace{0.02in}
\par \noindent \textbf{Replacing the decoder.}
In~\cref{tab:components_decoder_loss}, we observe that replacing our Transformer-based decoders by Box or Vote decoders leads to poor detection performance on both benchmarks.
Additionally, the Box and Vote decoders work only with their respective losses and our preliminary experiments using set loss on these decoders led to worse results.
Thus, the drop of performance could be attributed to changing the decoder used with our transformer encoder. We inspect this next by replacing the loss in \OURS while using the transformer encoder and decoder.

\vspace{0.02in}
\par \noindent \textbf{Replacing the loss.}
We train \OURS, \ie, both Transformer encoder and decoder with the Box and Vote losses.
We observe (\cref{tab:components_decoder_loss} rows \rownumber{4} and \rownumber{5}) that this leads to similar degradation in performance, suggesting that the losses are not applicable to our model.
This is not surprising since the design decisions, \eg, voting radius, aggregation radius \etc in the Vote loss was specifically designed for radius parameters in the PointNet++ encoder~\cite{qi2017pointnet++}.
This set of observations exposes that the decoder and loss function used in VoteNet depend greatly on the nature of the encoder (additional results in~\cref{app:masked_encoder_vote_loss}).
In contrast, our set loss has no design decisions specific to our encoder-decoder.

\begin{table}[!t]
    \mbox{}\hfill
      \centering
\setlength{\tabcolsep}{0.3em}\resizebox{\linewidth}{!}{
    \begin{threeparttable}
    \begin{tabular}{@{}ll|lll|cc|cc@{}}
    \toprule
    \rownumber{\#} & \textbf{Method} & \textbf{Encoder} & \textbf{Decoder} & \textbf{Loss}  & \multicolumn{2}{c}{\textbf{\scannet}} & \multicolumn{2}{c}{\textbf{\sunrgbd}}\\
    &&&&& AP$_{25}$ & AP$_{50}$ & AP$_{25}$ & AP$_{50}$ \\
        \hline
    \multicolumn{8}{l}{\textit{Comparing different decoders}} \\
    \hline
        \colorrow \rownumber{1}&\OURS & Tx. & Tx. & Set & 62.7 & 37.5 & 58.0 & 30.3 \\
    \rownumber{2}&& Tx. & Box  & Box & 31.0 & 10.2 & 36.4 & 14.4 \\
    \rownumber{3}&& Tx. & Vote  & Vote  & 46.1 & 23.4 & 47.5 & 24.9 \\

     \hline
    \multicolumn{8}{l}{\textit{Comparing different losses}} \\
    \hline
    \rownumber{4}&& Tx. & Tx.  & Box & 49.6 & 20.5 & 49.5 & 21.1 \\
    \rownumber{5}&& Tx. & Tx.  & Vote & 54.0  & 31.9 & 53.4 & 28.3 \\
            \bottomrule
    \end{tabular}
    \begin{tablenotes}
    \item {\small \hspace{-0.25in} Tx.: Transformer, Vote/Box loss~\cite{qi2019votenet}, Set loss~\cref{sec:set_loss} }
    \end{tablenotes}
    \end{threeparttable}
}
\vspace{-0.1in}
\caption{\textbf{\OURS with different decoders and losses.}
We vary the decoder and losses used with our transformer encoder.
As the Box and Vote decoders are only compatible with their losses, we vary the loss function while using them.
The Vote loss is compatible with our Transformer encoder-decoder, however a simpler set loss performs the best.
}
\label{tab:components_decoder_loss}
\end{table}

\vspace{0.02in}
\par \noindent \textbf{Visualizing self-attention.}
We visualize the self-attention in the decoder in~\cref{fig:encoder_attn}.
The decoder focuses on whole instances and groups points within instances.
This presumably makes it easier to predict bounding boxes for each instance.
We provide visualizations for the encoder self-attention in the supplemental.

\vspace{0.05in}
\par \noindent \textbf{Encoder applied to Shape classification.}
To verify that our encoder design is not specific to the detection task we test the encoder on shape classification of of models including 3D Warehouse~\cite{wu20153d}.

We use the three layer encoder from \OURS with vanilla self-attention (no decoder) or the three layer encoder from \OURSm.
To obtain global features for the point cloud, we use the `\texttt{CLS} token' formulation from Transformer, \ie, append a constant point to the input and use this point's output encoder features as global features (see supplemental for details).
The global features from the encoder are input to a 2-layer MLP to perform shape classification.
\cref{tab:shape_cls} shows that both the \OURS and \OURSm encoders are competitive with state-of-the-art encoders tailored for 3D.
These results suggest that our encoder design is not specific to detection and can be used for other 3D tasks.

\begin{table}[!t]
    \centering
    \setlength{\tabcolsep}{0.4em}
        \begin{tabular}{@{}l|ccc@{}}
        \toprule
        {\bf Method} & {\bf input} & {\bf mAcc} & {\bf OA} \\
        \hline

        \hline
        PointNet++~\cite{qi2017pointnet++} & point & -- & 91.9 \\
        SpecGCN~\cite{wang2018local} & point  & -- & 92.1 \\          DGCNN~\cite{wang2019dynamic} & point & 90.2 & 92.2 \\
        PointWeb~\cite{zhao2019pointweb} & point  & 89.4 & 92.3 \\
        SpiderCNN~\cite{xu2018spidercnn} & point  & -- & 92.4 \\
        PointConv~\cite{wu2019pointconv} & point  & -- & 92.5 \\
        KPConv~\cite{thomas2019kpconv} & point  & -- & 92.9 \\
        InterpCNN~\cite{mao2019interpolated} & point & -- & 93.0 \\
                \hline
                \colorrow \OURS encoder (Ours) & point & 89.1 & 92.1 \\          \colorrow \OURSm encoder (Ours) & point & 89.9 & 91.9 \\          \bottomrule
    \end{tabular}
        \vspace{-0.1in}
    \caption{\textbf{Shape classification.} We report shape classification results by training our Transformer encoder model.
    Our model performs competitively with architectures designed for 3D suggesting that our design decisions can extend beyond detection and be useful for other tasks.
    }
\label{tab:shape_cls}
\end{table}

\vspace{-0.1in}
\subsubsection{Design decisions in \OURS}
\label{sec:design_transformer}
\vspace{-0.1in}

\begin{table}[!t]
    \centering
\setlength{\tabcolsep}{0.4em}\resizebox{\linewidth}{!}{
    \begin{threeparttable}
    \begin{tabular}{@{}llllc|cc@{}}
    \toprule
    \rownumber{\#}&\textbf{Method} & \multicolumn{2}{c}{\textbf{Positional Embedding}} & \textbf{Query Type} & \multicolumn{2}{c}{\textbf{\scannet}}  \\      && Encoder& Decoder& & AP$_{25}$ & AP$_{50}$ \\         \hline
    \colorrow \rownumber{1} &\OURS & - & Fourier &  np + Fourier & 62.7 & 37.5 \\
    \rownumber{2}&& Fourier & Fourier &  np + Fourier & 61.8 & 37.0 \\
    \rownumber{3}&& Sine & Sine &  np + Sine & 55.8 & 30.9 \\
            \rownumber{4}&& - & - &  np + Sine & 31.3 & 10.8 \\
    \hline
    \rownumber{5} & DETR~\cite{carion2020end}$^\dagger$ & Sine & Sine & parametric~\cite{carion2020end} & 15.4 & 5.3 \\
    \bottomrule
    \end{tabular}
    \begin{tablenotes}
    \item {\small \hspace{-0.2in} np: non-parametric query (~\cref{sec:pointdetr})}
    \end{tablenotes}
    \end{threeparttable}
}
\vspace{-0.1in}
\caption{\textbf{Decoder Query Type and Positional Embedding.} We how using non-parametric queries and Fourier positional embeddings~\cite{tancik2020fourfeat} affect detection performance.
DETR's parametric queries do not work well for 3D detection (rows \rownumber{3}, \rownumber{5}).
The standard choice~\cite{vaswani2017attention,carion2020end} of sinusoidal positional embeddings is worse than Fourier embeddings (rows \rownumber{2}, \rownumber{3}).
$^\dagger$ - DETR is designed for 2D image detection and we adapt it for 3D detection.
}
\label{tab:ablate_query_position}
\end{table}

Our model is inspired by the DETR~\cite{carion2020end} architecture but has major differences - (1) it is an end-to-end transformer without a ConvNet, (2) it is trained from scratch (3) uses non-parametric queries and (4) Fourier positional embeddings.
In~\cref{tab:ablate_query_position}, we show the impact of the last two differences by evaluating various versions of our model on \scannet.
The version with minimal modifications is a DETR model applied to 3D with our training and loss function.

First, this version does not perform well on the \scannet benchmark, achieving 15\% AP$_{25}$.
However, when replacing the parametric queries by non-parametric queries, we observe a significant improvement of +40\% in AP$_{25}$ (\cref{tab:ablate_query_position} rows \rownumber{3} and \rownumber{5}).
In fact, only using the non-parametric queries (row \rownumber{4}) without positional embeddings doubles the performance.
This shows the importance of using non-parametric queries with 3D point clouds.
A reason is that point clouds are irregular and sparse, making the learning of parametric queries harder than on a 2D image grids.
Non-parametric queries are directly sampled from the point clouds and hence are less impacted by these irregularities.
Unlike the fixed number of parametric queries in DETR, non-parametric queries easily enable the use different number of queries at train and test time (see~\cref{sec:adaptive}).

Finally, replacing the sinusoidal positional embedding by the low-frequency Fourier encodings of~\cite{tancik2020fourfeat} provides an additional improvement of +5\% in AP$_{25}$ (\cref{tab:ablate_query_position} rows \rownumber{2} and \rownumber{3}).
As a side note, using positional encodings benefits the decoder more than the encoder because the decoder does not have direct access to coordinates.

\begin{table}[!t]
    \centering
    \begin{tabular}{@{}l|cc@{}}
    \toprule
    \textbf{Method} & NMS & No NMS \\
    \hline
    VoteNet~\cite{qi2019votenet} & 60.4 & 10.7 \\
    \OURS (ours) & 62.7 & 59.5 \\
    \bottomrule
    \end{tabular}
\vspace{-0.1in}
\caption{\textbf{Effect of NMS.} We report the detection performance (AP$_{25}$) for \OURS and VoteNet on \scannet.
\OURS works without NMS at test time because the set matching loss discourages excess predicted boxes.
}
\label{tab:ablate_nms}
\end{table}

\vspace{-0.05in}
\section{Ablations}
\label{sec:ablations}
\vspace{-0.05in}

We conduct a series of ablation experiments to understand the components of \OURS with settings from~\cref{sec:experiments}.

\begin{figure}[!t]
\includegraphics[width=0.45\textwidth]{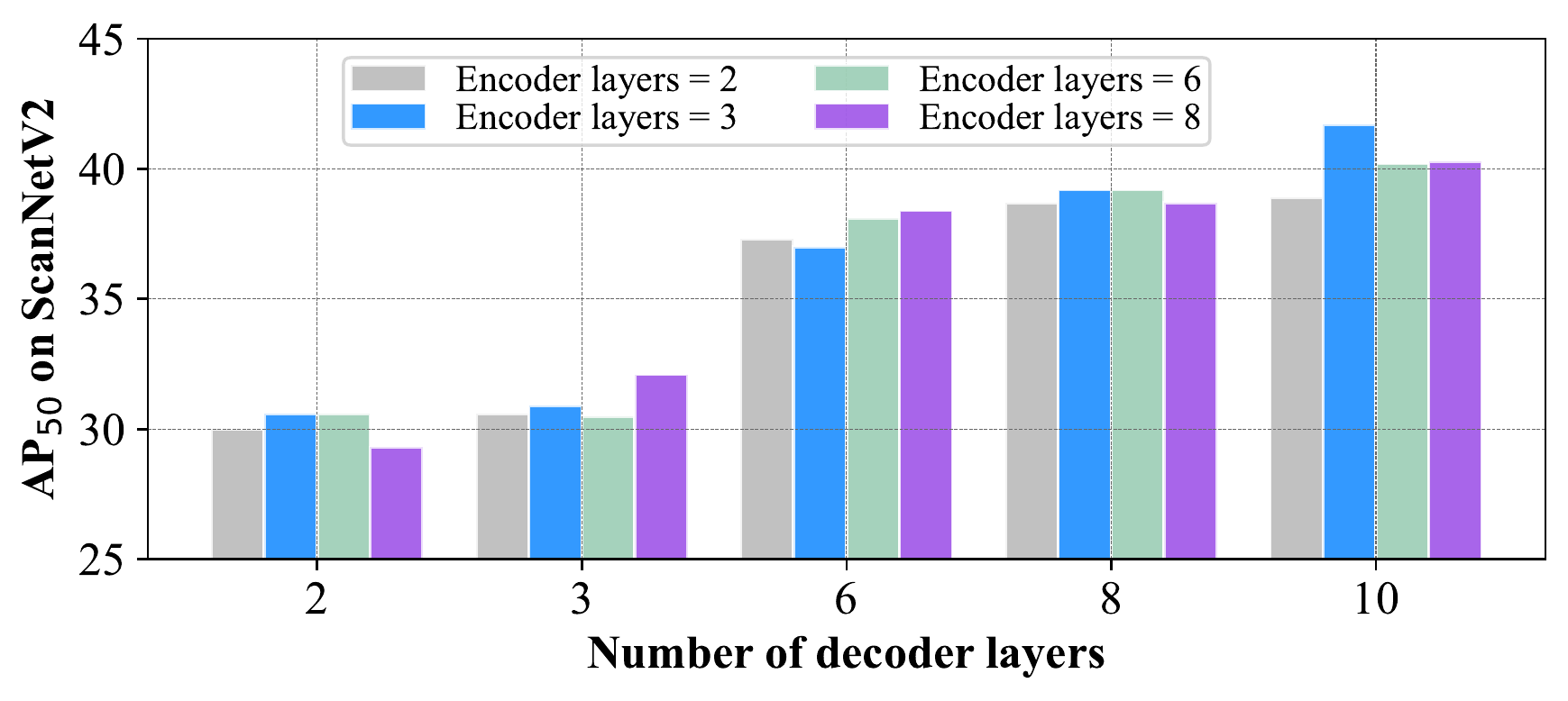}
\vspace{-0.2in}
\caption{\textbf{Varying number of layers for encoder and decoder.} 
We train different models with varying number of encoder and decoder layers and analyze the impact on detection performance on \scannet.
Increasing the number of layers in either the encoder or decoder has a positive effect, but a higher number of decoder layers matters more than the encoder layers.
}
\label{fig:enc_dec_layers}
\end{figure}

\begin{figure}[!t]
\centering

\includegraphics[width=0.23\textwidth]{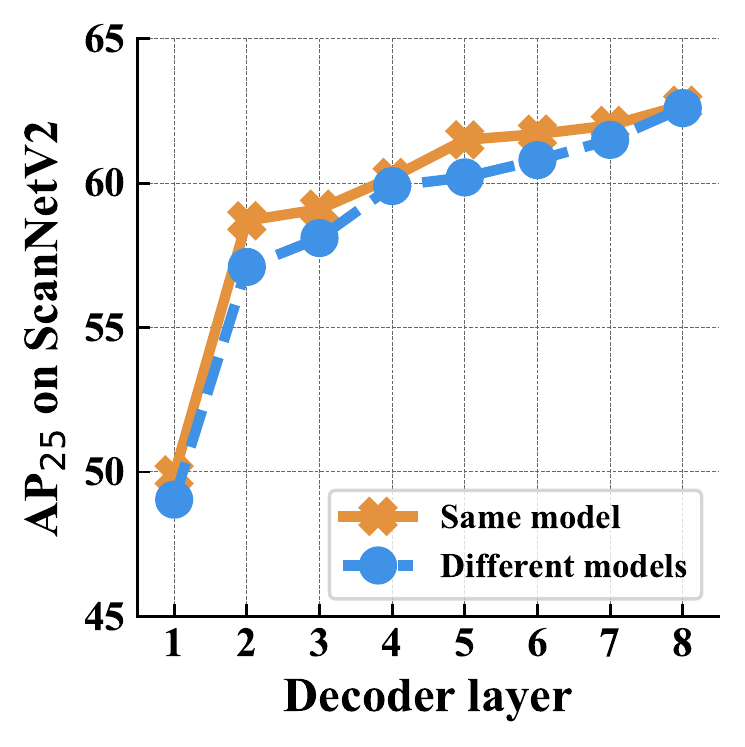}
\includegraphics[width=0.23\textwidth]{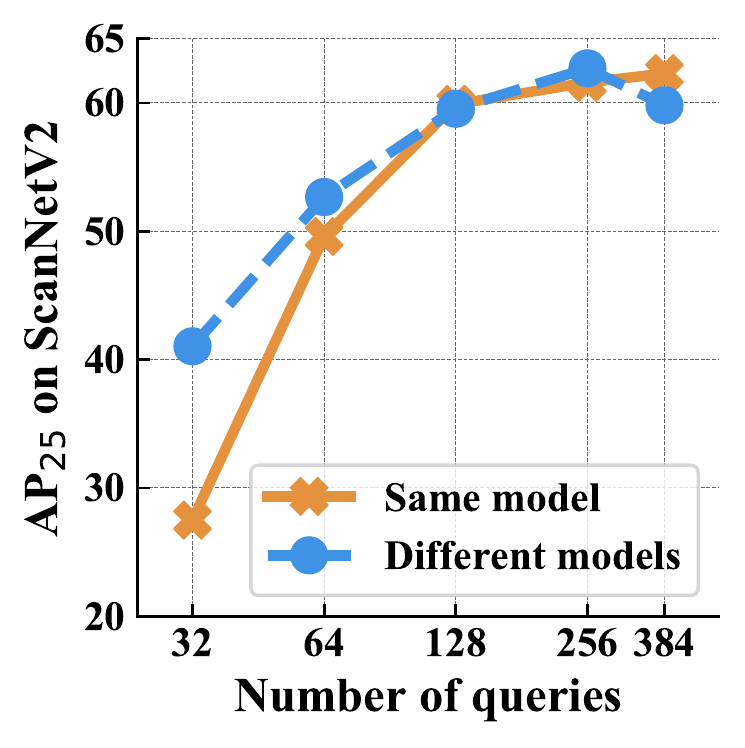}
\vspace{-0.15in}
\caption{\textbf{Adapting compute at test time.} 
We change the number of decoder layers or the number of queries used at test time for a \OURS model (`same model').
We compare this to different models trained with reduced depth of the decoder (left) or with different number of queries (right).
\OURS can adapt to different test time conditions and performs favorably compared to different models trained for the test time conditions.
}

\label{fig:dec_interim_queries}
\end{figure}

\vspace{0.05in}
\par \noindent \textbf{Effect of NMS.} 
\OURS uses the set loss of DETR (\cref{sec:set_loss}) that forces a 1-to-1 mapping between the ground truth box and the predicted box.
This loss penalizes models that predict too many boxes, since excess predictions are not matched to ground truth.
In contrast, the loss used in VoteNet~\cite{qi2019votenet} does not discourage multiple predictions of the same object and thus relies on Non-Maximal Suppression to remove them as a post-processing step.
We compare \OURS and VoteNet with and without NMS in~\cref{tab:ablate_nms} with the detection AP metric, which penalizes duplicate detections.
Without NMS, \OURS drops in performance by only 3\% AP while VoteNet drops by 50\%, showing our set loss works without NMS.

\vspace{0.06in}
\par \noindent \textbf{Effect of encoder/decoder layers.}
We assess the importance of the number of layers in the encoder and decoder in~\cref{fig:enc_dec_layers}.
While a higher number of layers improves detection performance in general, adding the layers in the decoder instead of the encoder has a greater impact on performance.
For instance, for a model with three encoder and three decoder layers, adding five decoder layers improves performance by +7\% AP$_{50}$ while adding five encoder layers improves by +2\%AP$_{50}$.
This preference toward the decoder arises because in our parallel decoder, each layer further refines the prediction quality of the bounding boxes.

\subsection{Adapting computation to inference constraints}
\label{sec:adaptive}
An advantage of our model is that we can adapt its computation during inference by using less layers in the decoder or queries to predict boxes without retraining.

\vspace{0.06in}
\par \noindent \textbf{Adapting decoder depth.} 
The parallel decoder of \OURS is trained to predict boxes at each layer with the same bounding box prediction MLPs.
Thus far, in all our results we used the predictions only from the last decoder layer.
We now test the performance of the intermediate layers for a decoder with six layers in~\cref{fig:dec_interim_queries} (left).
We compare this to training different models with a varying number of decoder layers.
We make two observations - (1) similar to~\cref{fig:enc_dec_layers}, detection performance increases with the number of decoder layers; and (2) more importantly, the same model with reduced depth at test time performs as well or better than models trained from scratch with reduced depth.
This second property is shared with the DETR, but not with VoteNet.
It allows adapting the number of layers in the decoder to a computation budget during inference without retraining.

\vspace{0.06in}
\par \noindent \textbf{Adapting number of queries.}
As we increase the number of queries, \OURS predicts more bounding boxes, resulting in better performance at a cost of longer running time.
However, our non-parametric queries in \OURS allow us to adapt the number of box predictions to trade performance for running time. Note that this is also possible with VoteNet, but not with DETR.
In~\cref{fig:dec_interim_queries} (right), we compare changing the number of queries at test time to different models trained with varying number of queries.
The same \OURS model can adapt to a varying number of queries at test time and performs comparably to different models.
Performance increases until the number of queries is enough to cover the point cloud well.
We found this adaptation to number of queries at test time works best with a \OURS model trained with $128$ queries (see~\cref{sec:supp_experiments} for other models).
This adaptive computation is promising and research into efficient self-attention should benefit our model.
We provide inference time comparisons to VoteNet in~\cref{sec:supp_architecture} for different versions of the \OURS model.

\vspace{-0.05in}
\section{Conclusion}
\label{sec:conclusion}
\vspace{-0.05in}

We presented \OURS, an end-to-end Transformer model for 3D detection on point clouds.
\OURS requires few 3D specific design decisions or hyperparameters.\
We show that using non-parametric queries and Fourier encodings is critical for good 3D detection performance.
Our proposed design decisions enable powerful Transformers for 3D detection, and also benefit other 3D tasks like shape classification.
Additionally, our set loss function generalizes to prior 3D architectures.
In general, \OURS is a flexible framework and can easily incorporate prior components used in 3D detection and can be leveraged to build more advanced 3D detectors.
Finally, it also combines the flexibility of both VoteNet and DETR, allowing for a variable number of predictions at test time (like VoteNet) with a variable number of decoder layers (like DETR).
\vspace{0.05in}
{\small
\par \noindent \textbf{Acknowledgments:} We thank Zaiwei Zhang for helpful discussions and Laurens van der Maaten for feedback on the paper.
}

\clearpage
\appendix
\section*{Supplemental Material}

\section{Implementation Details}

\subsection{Architecture}
\label{sec:supp_architecture}

We describe the \OURS architecture in detail.

\begin{figure}[!b]
\centering
\includegraphics[width=0.5\textwidth]{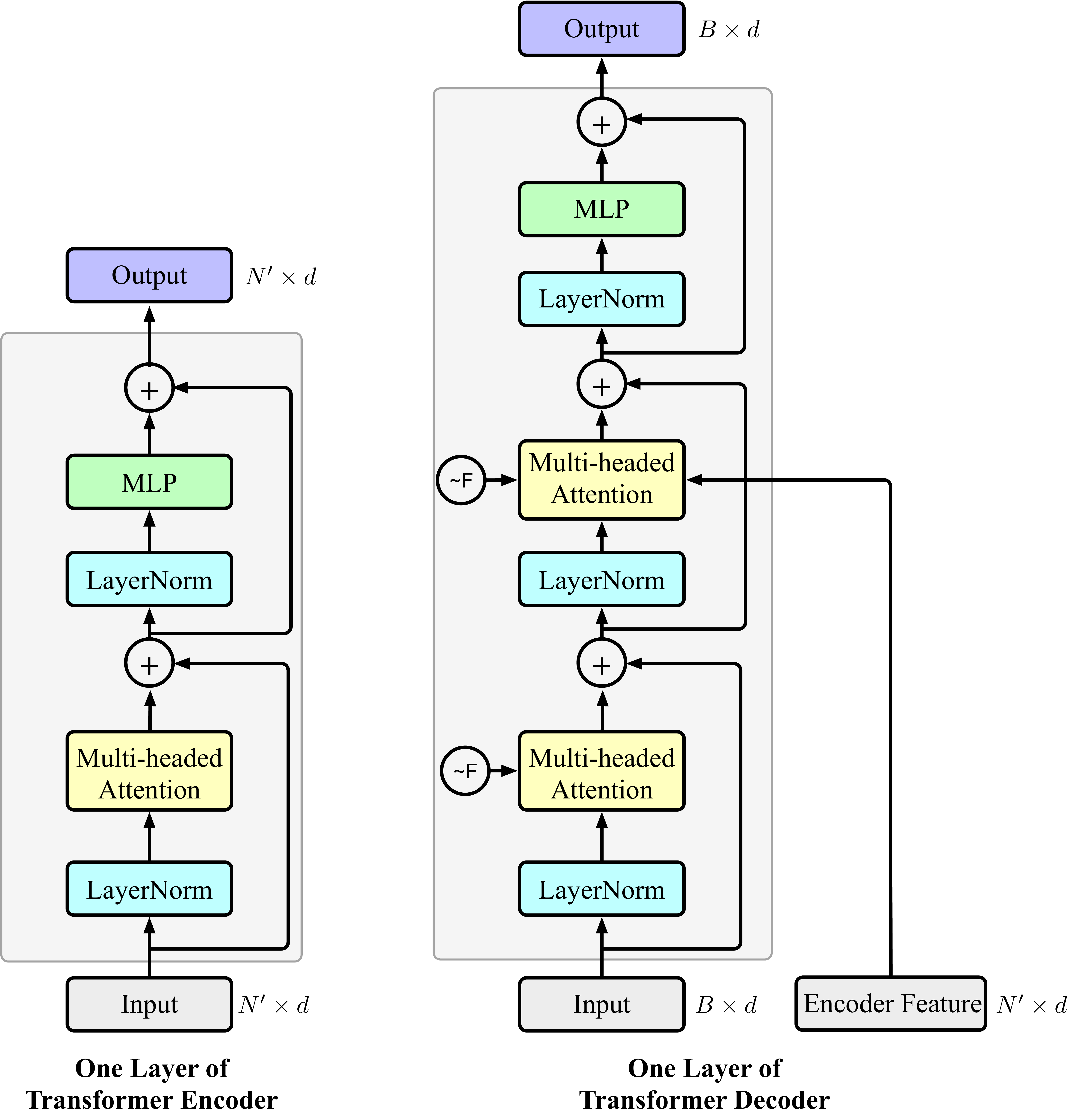}
\caption{\textbf{Architecture of Encoder and Decoder.}
We present the architecture for one layer of the \OURS encoder and decoder.
The encoder layer takes as input $\pN\times d$ features for $\pN$ points and outputs $\pN\times d$ features too.
It performs self-attention followed by a MLP.
The decoder takes as input $B\times d$ features (the query embeddings or the prior decoder layer), $\pN\times d$ point features from the encoder to output $B\times d$ features for $B$ boxes.
The decoder performs self-attention between the $B$ query/box features and cross-attention between the $B$ query/box features and the $\pN$ point features.
We denote by $\sim_{F}$ the Fourier positional encodings~\cite{tancik2020fourfeat} used in the decoder.
All \OURS models use $d=256$.
\label{fig:enc_dec_arch}
}
\end{figure}

\par \noindent \textbf{Architecture.}
We follow the dataset preprocessing from~\cite{qi2019votenet} and obtain $N=20,000$ points and $N=40,000$ points respectively for each sample in \sunrgbd and \scannet datasets.
The $N\times3$ matrix of point coordinates is then passed through one layer of the downsampling and set aggregation operation~\cite{qi2017pointnet++} which uses Farthest-Point-Sampling to sample $2048$ points randomly from the scene.
Each point is projected to a $256$ dimensional feature followed by the set-aggregation operation that aggregates features within a $\ell_{2}$ distance of $0.2$.
The output is a $2048\times256$ dimensional matrix of features for the $\pN=2048$ points which is input to the encoder.
We now describe the encoder and decoder architectures (illustrated in~\cref{fig:enc_dec_arch}).

\par \noindent \textbf{Encoder.}
The encoder has three layers of self-attention followed by an MLP.
The self-attention operation uses multi-headed attention with four heads.
The self-attention produces a $2048\times2048$ attention matrix which is used to attend to the  features to produce a $256$ dimensional output.
The MLPs in each layer have a hidden dimension with $128$.
All the layers use LayerNorm~\cite{ba2016layer} and the ReLU non-linearity.

\par \noindent \textbf{\OURSm Encoder.}
The masked \OURSm encoder has three layers of self-attention followed by an MLP.
At each layer the self-attention matrix of size \texttt{\#points}$\times$\texttt{\#points} is multiplied with a binary mask $M$ of the same size.
The binary mask entry $M_{ij}$ is $1$ if the point coordinates for points $i$ and $j$ are within a radius $r$ of each other.
We use radius values of $[0.4, 0.8, 1.2]$ for the three layers.
The first layer operates on $2048$ points and is followed by a downsample + set aggregation operator that downsamples to $1024$ points using a radius of $0.4$, similar to PointNet++.
The encoder layers follow the same structure as the vanilla Encoder described above, \ie, MLPs with hidden dimension of $128$, multi-headed attention with four heads \etc.
The encoder produces $256$ dimensional features for $1024$ points.

\par \noindent \textbf{Decoder.}
The decoder operates on the $\pN\times256$ encoder features and $B\times256$ location query embeddings.
It produces a $B\times256$ matrix of box features as output.
The decoder has eight layers and uses cross-attention between the location query embeddings (Sec 3.2 main paper) and the encoder features, and self-attention between the box features.
Each layer has the self-attention operation followed by a cross-attention operation (implemented exactly as self-attention) and an MLP with a hidden dimension of $256$.
All the layers use LayerNorm~\cite{ba2016layer}, ReLU non-linearity and a dropout of $0.3$.

\par \noindent \textbf{Bounding box prediction MLPs.}
The box prediction MLPs operate on the $B\times256$ box features from the decoder.
We use separate MLPs for the following five predictions - 1) center location offset $\Delta \bq \in [0,1]^{3}$; 2) angle quantization class; 3) angle quantization residual $\in \mathbb{R}$; 4) box size $\bs \in [0,1]^{3}$; 5) semantic class of the object.
Each MLP has $256$ hidden dimensions and uses the ReLU non-linearity.
The center location and size prediction MLP outputs are followed by a sigmoid function to convert them into a $[0,1]$ range.

\par \noindent \textbf{Inference speed}.
\OURS has very few 3D-specific tweaks and uses standard PyTorch.
VoteNet relies on custom GPU CUDA kernels for 3D operations.
We measured the inference time of \OURS (256 queries) and VoteNet (256 boxes) on a V100 GPU with a batchsize of 8 samples.
Both models downsample the pointcloud to $2048$ points.
\OURS needs 170 ms while VoteNet needs 132 ms.
As research into efficient self-attention becomes more mature (several recent works show promise), it will benefit the runtime and memory efficiency of our model.

\begin{table}[!t]
    \centering
  \setlength{\tabcolsep}{0.3em}\resizebox{\linewidth}{!}{
    \begin{tabular}{@{}cc|c@{}}
    \toprule
    \textbf{Encoder Layers} & \textbf{Decoder Layers} & \textbf{Inference time} \\
    \hline
    3 & 3 & 153 \\
    3 & 6 & 164 \\
    3 & 8 & 170 \\
    3 & 10 & 180 \\
    \hline
    6 & 6 & 193 \\
    6 & 8 & 213 \\
    \hline
    8 & 8 & 219 \\
    \bottomrule
    \end{tabular}
}
\vspace{-0.1in}
\caption{\textbf{Inference Speed and Memory.}
We provide inference speed (in milliseconds) for different number of encoder and decoder layers in the \OURS model.
All timings are measured on a single V100 GPU with a batchsize of 8 and using 256 queries.
}
\label{tab:inference_speed}
\end{table}

\subsection{Set Loss}
The set matching cost is defined as:
\begin{equation*}
  C_{\bmatch}(\hbb, \bb) =  \underbrace{- \lambda_{1} \mathrm{GIoU}(\hbb, \bb) + \lambda_{2} \|\hbc - \bc\|_{1}}_\text{geometric} -\underbrace{\lambda_{3} \hbs[s_{gt}]}_\text{semantic} \\
\end{equation*}

For $B$ predicted boxes and $G$ ground truth boxes, we compute a $B\times G$ matrix of costs by using the above pairwise cost term.
We then compute an optimal assignment between each ground truth box and predicted box using the Hungarian algorithm.
Since the number of predicted boxes is larger than the number of ground truth boxes, the remainder $B - G$ boxes are considered to match to background.
We set $\lambda_{1}, \lambda_{2}, \lambda_{3}, \lambda_{4}$ as $2, 1, 0, 0$ for \scannet and $3,5,1, 5$ for \sunrgbd.

For each predicted box that is matched to a ground truth box, our loss function is:
\begin{multline*}
\mathcal{L}_{\mathrm{\OURS}} = 5 * \|\hbc - \bc\|_{1} + \|\hbd - \bd\|_{1} + \|\hba_r - \ba_r\|_{\mathrm{huber}} \\
-0.1 * \ba_c^\intercal \log \hba_c - 5 * \bs_c^\intercal \log \hbs_c
\end{multline*}
For each unmatched box that is considered background, we compute only the semantic loss term.
The semantic loss is implemented as a weighted cross entropy loss with the weight of the `background' class as $0.2$ and a weight of $0.8$ for the K object classes.

\section{Experiments}
\label{sec:supp_experiments}
We provide additional experimental details and hyperparameter settings.

\subsection{Improved baselines}
We improve the VoteNet and BoxNet baselines by doing a grid search and improving the optimization hyperparameters.
We train the baseline models for $360$ epochs using the Adam optimizer~\cite{KingmaB14} with a learning rate of $1\times10^{-3}$ decayed by a factor of 10 after $160, 240, 320$ epochs and a weight decay of $0$.
We found that using a cosine learning rate schedule, even longer training than 360 epochs or the AdamW optimizer did not make a significant difference in performance for the baselines.
These improvements to the baseline lead to an increase in performance, summarized in~\cref{tab:improved_baseline}.

\begin{table}[!t]
      \centering
    \begin{tabular}{@{}l|cccc@{}}
    \toprule
    \textbf{Method} & \multicolumn{2}{c}{\textbf{\scannet}} & \multicolumn{2}{c}{\textbf{\sunrgbd}} \\
    &AP$_{25}$ & AP$_{50}$ & AP$_{25}$ & AP$_{50}$ \\
    \hline
    BoxNet~\cite{qi2019votenet} & 45.4 & - & 53.0 & - \\
    BoxNet$^\dagger$~\cite{qi2019votenet} & 49.0 & 21.1 & 52.4 & 25.1 \\
    \hline
    VoteNet~\cite{qi2019votenet} & 58.6 & 33.5 & 57.7 & - \\
    VoteNet$^\dagger$~\cite{qi2019votenet} & 60.4 & 35.5 & 58.3 & 33.4 \\
    \bottomrule
    \end{tabular}
\vspace{-0.1in}
\caption{\textbf{Improved baseline.}
We denote by $^\dagger$ our improved implementation of the baseline methods and report the numbers from the original paper~\cite{qi2019votenet}.
Our improvements ensure that the comparisons in the main paper are fair.
}
\label{tab:improved_baseline}
\end{table}

\subsection{Per-class Results}
We provide the per-class mAP results for \scannet in~\cref{tab:per_class_scannet} and \sunrgbd in~\cref{tab:per_class_sun}.
The overall results for these models were reported in the main paper (~\cref{tab:comparison}).

\begin{table*}[!t]
  \centering
  \begin{tabular}{@{}l|cccccccccc@{}}
    \toprule
    Model & bed & table & sofa & chair & toilet & desk & dresser & nightstand & bookshelf & bathtub\\
    \hline
    \OURS & 81.8 & 50.0 & 58.3 & 68.0 & 90.3 & 28.7 & 28.6 & 56.6 & 27.5 & 77.6 \\
    \OURS-m & 84.6 & 52.6 & 65.3 & 72.4 & 91.0 & 34.3 & 29.6 & 61.4 & 28.5 & 69.8 \\
  \bottomrule
  \end{tabular}
\vspace{-0.1in}
\caption{\textbf{Per-class AP$_{25}$ for \sunrgbd.}
}
\label{tab:per_class_sun}
\end{table*}

\begin{table*}[!t]
  \centering
\setlength{\tabcolsep}{0.3em}\resizebox{\linewidth}{!}{
  \begin{tabular}{@{}l|cccccccccccccccccc@{}}
    \toprule
    Model & cabinet & bed & chair & sofa & table & door & window & bookshelf & picture & counter & desk & curtain & refrigerator & showercurtrain & toilet & sink & bathtub & garbagebin \\
    \hline
    \OURS & 50.2 & 87.0 & 86.0 & 87.1 & 61.6 & 46.6 & 40.1 & 54.5 & 9.1 & 62.8 & 69.5 & 48.4 & 50.9 & 68.4 & 97.9 & 67.6 & 85.9 & 45.8 \\
    \OURSm & 49.4 & 83.6 & 90.9 & 89.8 & 67.6 & 52.4 & 39.6 & 56.4 & 15.2 & 55.9 & 79.2 & 58.3 & 57.6 & 67.6 & 97.2 & 70.6 & 92.2 & 53.0 \\
  \bottomrule
  \end{tabular}
}
\vspace{-0.1in}
\caption{\textbf{Per-class AP$_{25}$ for \scannet.}
}
\label{tab:per_class_scannet}
\end{table*}

\subsection{Detailed state-of-the-art comparison}
We provide a detailed comparison to state-of-the-art detection methods in~\cref{tab:sota_comparison}.
Most state-of-the-art methods build upon VoteNet.
H3DNet~\cite{zhang2020h3dnet} uses 3D primitives with VoteNet for better localization.
HGNet~\cite{chen2020hierarchical} improves VoteNet by using a hierarchical graph network with higher resolution output from its PointNet++ backbone.
3D-MPA~\cite{engelmann20203d} uses clustering based geometric aggregation and graph convolutions on top of the VoteNet method.
\OURS does not use Voting and has fewer 3D specific decisions compared to all other methods.
\OURS performs favorably compared to these methods and outperforms VoteNet.
This suggests that, like VoteNet, \OURS can be used as a building block for future 3D detection methods.

\begin{table}[!t]
          \centering
\setlength{\tabcolsep}{0.3em}\resizebox{\linewidth}{!}{
    \begin{tabular}{@{}l|lcccc@{}}
    \toprule
    \textbf{Method} & \textbf{Arch.} & \multicolumn{2}{c}{\textbf{\scannet}} & \multicolumn{2}{c}{\textbf{\sunrgbd}}\\
    && AP$_{25}$ & AP$_{50}$ & AP$_{25}$ & AP$_{50}$ \\
    \hline
    BoxNet$^\dagger$~\cite{qi2019votenet} & BoxNet & 49.0 & 21.1 & 52.4 & 25.1 \\
    \OURS & Tx. & 62.7 & 37.5 & 56.8 & 30.1 \\
\hline
    VoteNet$^\dagger$~\cite{qi2019votenet} & VoteNet & 60.4 & 37.5 & 58.3 & 33.4 \\
    \OURSm & Tx. & 65.0 & 47.0 & 59.0 & 32.7 \\
\hline
    H3DNet~\cite{zhang2020h3dnet} & VoteNet + 3D primitives & 67.2 & 48.1 & 60.1 & 39.0 \\
    HGNet~\cite{chen2020hierarchical} & VoteNet + GraphConv & 61.3 & 34.4 & 61.6 & 34.4 \\
    3D-MPA~\cite{engelmann20203d} & VoteNet + GraphConv & 64.2 & 49.2 & - & - \\
  \bottomrule
    \end{tabular}
}
\vspace{-0.1in}
\caption{\textbf{Detailed state-of-the-art comparison on 3D detection.}
}
\label{tab:sota_comparison}
\end{table}

\subsection{\OURSm with Vote loss}
\label{app:masked_encoder_vote_loss}
We tuned the VoteNet loss with the \OURSm encoder and our best tuned model gave 60.7\% and 56.1\% mAP on \scannet and \sunrgbd respectively (settings from~\cref{tab:components_decoder_loss} of the main paper).
The VoteNet loss performs better with \OURSm compared to the vanilla \OURS encoder (gain of 6\% and 3\%), confirming that the VoteNet loss is dependent on the inductive biases/design of the encoder.
Using our set loss is still better than using the VoteNet loss for \OURSm (~\cref{tab:comparison} \vs results stated in this paragraph).
Thus, our set loss design decisions are more broadly applicable than that of VoteNet.

\subsection{Adapt queries at test time}
\begin{figure}
\includegraphics[width=0.23\textwidth]{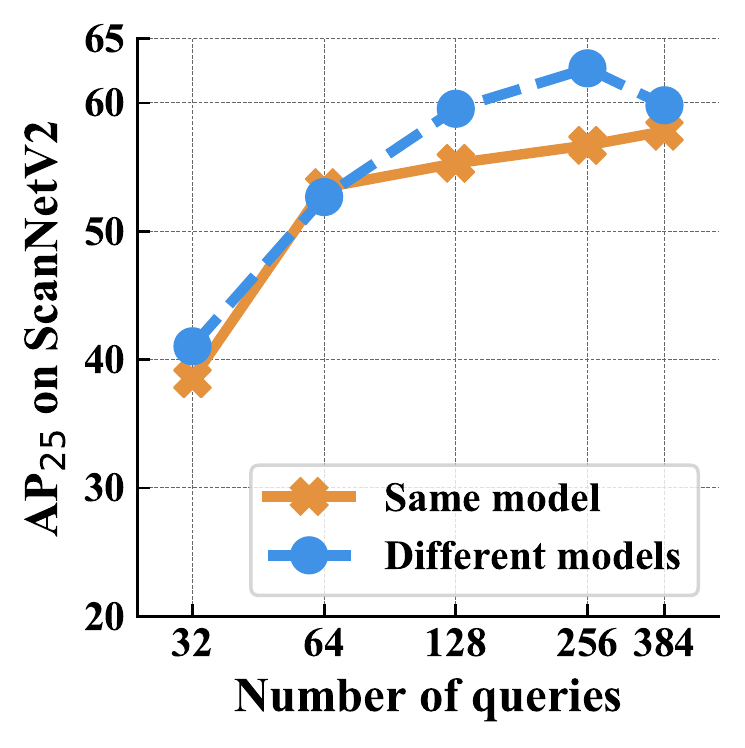}
\includegraphics[width=0.23\textwidth]{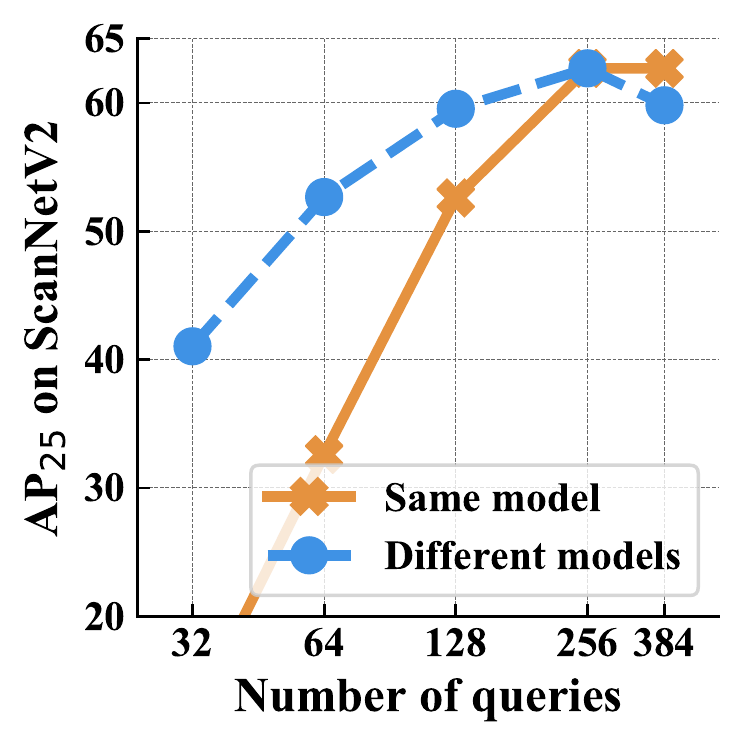}
\caption{\textbf{Adapt queries at test time.} Similar to Figure 5 of the main paper, we change the number of queries at test time for a \OURS model and compare it to different models trained with a varying number of queries.
We plot the results for the same \OURS model trained with 64 queries (left) or with 256 queries (right).
\label{fig:queries_test_time_extra}
}
\end{figure}

We provide additional results for Section 5.1 of the main paper.
We change the number of queries used at test time for the same \OURS model.
We show these results in~\cref{fig:queries_test_time_extra} for two different \OURS models trained with 64 and 256 queries respectively.
We observe that the model trained with $64$ queries is more robust to changing queries at test-time, but at its most optimal setting achieves worse detection performance than the model trained with $256$ queries.
In the main paper, we show results of changing queries at test time for a model trained with $128$ queries that achieves a good balance between overall performance and robustness to change at test-time.

\subsection{Visualizing the encoder attention}
\begin{figure}
\centering
\includegraphics[width=0.5\textwidth]{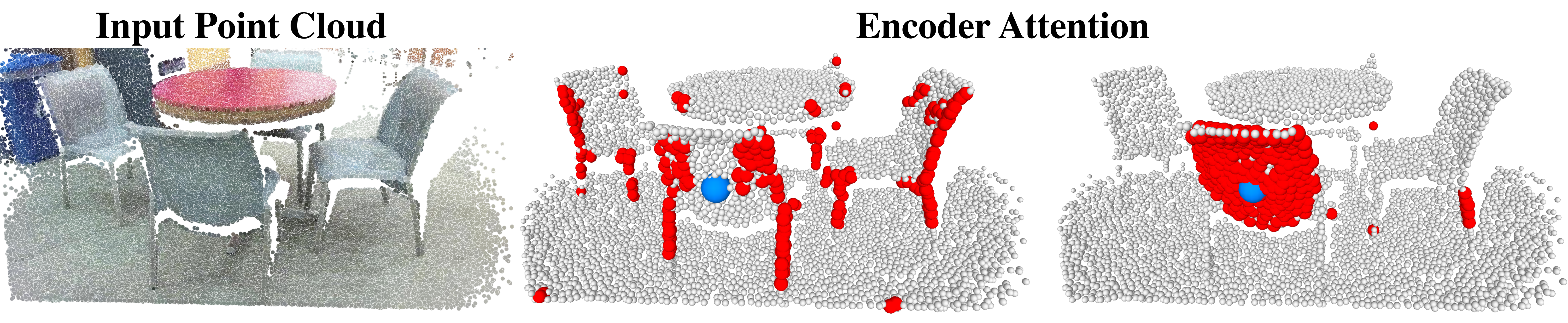}
\caption{\textbf{Encoder attention.} We visualize the encoder attention for two different heads.
We compute the self-attention from the reference point (blue dot) to all the points in the scene and display the points with the highest attention values in red.
The encoder groups together different geometric parts (legs of multiple chairs) or focuses on single parts of an instance (backrest of a chair).
\label{fig:encoder_attention}}
\end{figure}

We visualize the encoder attention for a \OURS model trained on the \sunrgbd dataset in~\cref{fig:encoder_attention}.
The encoder focuses on parts of objects.

\subsection{Shape Classification setup}

\paragraph{Dataset and Metrics.}
We use the processed point clouds with normals from~\cite{qi2017pointnet++}, and sample 8192 points as input for both training and testing our models.
Following prior work~\cite{zhao2019pointweb}, we report two metrics to evaluate shape classification performance: 1) Overall Accuracy (OA) evaluates how many point clouds we classify correctly; and 2) Class-Mean Accuracy (mAcc) evaluates the accuracy for each class independently, followed by an average over the per-class accuracy. This metric ensures tail classes contribute equally to the final performance.

\paragraph{Architecture Details.}
We use the base \OURS and \OURSm encoder architectures, followed by a 2-layer MLP with batch norm and a 0.5 dropout to transform the final features into a distribution over the 40 predefined shape classes. Differently from object detection experiments, our point features include the 3D position information concatenated with 3D normal information at each point, and hence the first linear layer is correspondingly larger, though the rest of the network follows the same architecture as the encoder used for detection. For the experiments with \OURS, we prepend a \texttt{[CLS]} token, output of which is used as input to the classification MLP. For the experiments with \OURSm that involve masked transformers, we max pool the final layer features, which are then passed into the classifier.

\paragraph{Training Details.} All models are trained for 250 epochs with a learning rate of $4\times10^{-4}$ and a weight decay of $0.1$, using the AdamW optimizer. We use a linear warmup from $4\times10^{-7}$ to the initial LR over 20 epochs, and then decay to $4\times10^{-5}$ over the remaining 230 epochs.
The models are trained on 4 GPUs with a batch size of 2 per GPU.

{\small
\bibliographystyle{ieee_fullname}
\bibliography{refs}
}

\end{document}